\documentclass[11pt,a4paper]{article}
\usepackage[hyperref]{emnlp2020}
\usepackage[utf8x]{inputenc}
\usepackage{times}
\usepackage{latexsym}
\usepackage{booktabs}
\usepackage{multirow}
\usepackage{url}
\usepackage{subcaption}
\usepackage{graphicx}

\usepackage{microtype}

\aclfinalcopy 

\title{PARADE: A New Dataset for Paraphrase Identification Requiring Computer Science Domain Knowledge}
\author{Yun He, Zhuoer Wang, Yin Zhang, Ruihong Huang, James Caverlee \\
  Department of Computer Science and Engineering, Texas A\&M University \\
  \texttt{\{yunhe, wang, zhan13679, huangrh, caverlee\}@tamu.edu} }

\date{}

\begin{document}
\maketitle
\begin{abstract}
We present a new benchmark dataset called PARADE for paraphrase identification that requires specialized domain knowledge. PARADE contains paraphrases that overlap very little at the lexical and syntactic level but are semantically equivalent based on computer science domain knowledge, as well as non-paraphrases that overlap greatly at the lexical and syntactic level but are not semantically equivalent based on this domain knowledge. Experiments show that both state-of-the-art neural models and non-expert human annotators have poor performance on PARADE. For example, BERT after fine-tuning achieves an F1 score of 0.709, which is much lower than its performance on other paraphrase identification datasets. PARADE can serve as a resource for researchers interested in testing models that incorporate domain knowledge. We make our data and code freely available.\footnote{\url{https://github.com/heyunh2015/PARADE_dataset}}

\end{abstract}


\section{Introduction}
\label{section: Introduction}
Paraphrases are sentences that express the same (or similar) meaning by using different wording \cite{bhagat2013paraphrase}. Automatically identifying paraphrases and non-paraphrases has proven useful for a wide range of natural language processing (NLP) applications, including question answering, semantic parsing, information extraction, machine translation, textual entailment, and semantic textual similarity.

Paraphrase identification (PI) is typically formalized as a binary classification problem: given two sentences, determine if they roughly express the same meaning. Traditional paraphrase identification approaches \cite{mihalcea2006corpus, kozareva2006paraphrase, wan2006using, das2009paraphrase, xu2014extracting} mainly rely on lexical and syntactic overlap 
features to measure the semantic similarity between the two sentences. Examples include 
string-based 
features (e.g., whether two sentences share the same words), part-of-speech features (e.g., whether shared words have the same POS tags), and dependency-based features (e.g., whether two sentences have similar dependency trees). 

\begin{table}[htbp]
  \centering
\small
\setlength{\tabcolsep}{0.7pt}
\renewcommand\arraystretch{1.0}
\newcommand{\tabincell}[2]{\begin{tabular}{@{}#1@{}}#2\end{tabular}}

    \begin{tabular}{l}
 \toprule
     s1: the lowest level of code made up of 0s and 1s. \\
           s2: binary instructions used by the cpu. \\
		Label: paraphrase (both describe ``Machine Code'') \\
 \midrule
     \tabincell{l}{s3: a graph representation that uses a 2d array such that if \\ \ arr[i][j] == 1, there is an edge between \textbf{vertices} i and j} \\
          \tabincell{l}{s4: a matrix which records the number of direct links between  \\ \ \textbf{vertices}} \\
 	Label: paraphrase  (both describe ``Adjacency Matrix'')\\

 
 \midrule
  \tabincell{l}{s5: how \textbf{the optimal solution} to \textbf{a linear programming} \\ \ \textbf{problem} changes as the \underline{problem data} are modified.}  \\
 
           \tabincell{l}{s6: how changes in the \underline{coefficients} of \textbf{a linear programming} \\ \ \textbf{problem}  affect \textbf{the optimal solution}}   \\
 Label: non-paraphrase \\
\bottomrule
    \end{tabular}%
\caption{Examples of paraphrases and non-paraphrases from the computer science domain. Judgments are made based on domain knowledge rather than lexical or syntactic features. Overlapping words (other than stop-words) are in bold and key different words are underlined.}
  \label{tab:Examples of paraphrases based on domain knowledge}%
\end{table}%

However, these shallow lexical and syntactic overlap 
features may not effectively capture the domain-specific semantics of the two sentences. A typical situation where models based on these overlap
features may fail is \textit{a pair of sentences that overlap very little at the lexical and syntactic level but are semantically equivalent \textbf{based on domain knowledge}}. Consider the two paraphrases s1 and s2 in Table \ref{tab:Examples of paraphrases based on domain knowledge}. Both describe \textit{machine code} though they have very little overlap. In order to correctly identify paraphrases like this pair, it is necessary to have specialized domain knowledge that a processor (CPU) can only understand binary instructions made up of 0s and 1s. On the other hand, \textit{a pair of sentences that overlap greatly at the lexical and syntactic level but are not semantically equivalent \textbf{based on domain knowledge}} can also confuse both non-expert annotators and NLP models. Consider the non-paraphrase of s5 and s6 in Table \ref{tab:Examples of paraphrases based on domain knowledge} as an example. Sentence s5 is about a sensitivity analysis between the problem data and the optimal solution while s6 is about a sensitivity analysis between the coefficients and the optimal solution; these two cases are fundamentally different, requiring specialized domain knowledge of linear programming to distinguish the two. These examples highlight the importance of \textbf{specialized domain knowledge} for identifying paraphrases and non-paraphrases correctly.

Recent neural models \cite{nie2017shortcut, parikh-etal-2016-decomposable, chen-etal-2017-enhanced} that go beyond traditional approaches based on lexical and syntactic features have demonstrated  state-of-the-art performance on paraphrase identification. For example, BERT and its variants \cite{devlin2018bert, liu2019roberta, yang2019xlnet, lan2019albert, raffel2019exploring} have achieved the best results on the General Language Understanding Evaluation (GLUE) benchmark \cite{wang2018glue} on two paraphrase identification datasets: the Microsoft Research Paraphrase Corpus (MRPC) and Quora Question Pairs (QQP). Using massive pre-training data and a flexible bidirectional self-attention mechanism, BERT and its variants are able to better model the semantic relationship between sentences. Moreover, two recent studies \cite{petroni-etal-2019-language, davison-etal-2019-commonsense} observe that BERT without fine-tuning can even capture world knowledge and can answer factual questions like ``place of birth'' and ``who developed the theory of relativity.'' Naturally, we are curious to know if these neural models can correctly identify paraphrases that require specialized domain knowledge like the examples shown in Table \ref{tab:Examples of paraphrases based on domain knowledge}.




Hence, our overarching research goal is to create new datasets and enable new models for high-quality  \textit{paraphrase identification based on domain knowledge}. Because previous paraphrase datasets \cite{dolan-brockett-2005-automatically, dolan-etal-2004-unsupervised, xu2014extracting, lan2017continuously, iyer2017first, zhang-etal-2019-paws} were not originally designed and constructed from the perspective of domain knowledge, to date there is no such dataset that requires specialized domain knowledge to discern the quality of two candidate sentences as paraphrases. As a first step, we focus in this paper on the computer science domain. Specifically, we require a dataset of paraphrases that overlap very little but are semantically equivalent, and of non-paraphrases that have overlap greatly but are not semantically equivalent based on computer science domain knowledge. Correspondingly, there is a research gap in understanding if modern neural models can achieve exemplary performance on such a dataset, especially in comparison with existing paraphrase identification datasets (that lack such specialized domain knowledge). In sum, this paper makes four contributions:






\begin{itemize}
	\item First, we propose a novel extensible framework for inexpensively collecting domain-specific sentential candidate paraphrases that are characterized by specialized knowledge. The key idea is to leverage large-scale online collections of \textit{user-generated flashcards}. We treat definitions on each flashcard's back side that correspond to a common entity on the front side (e.g., ``machine code'') as candidate paraphrases. 


	\item Due to the noise in user-generated flashcards and heterogeneity in the aspects in the candidate paraphrases, our second contribution is a refinement strategy coupled with annotation by domain experts to create a new gold dataset called PARADE (\textbf{PARA}phrase identification based on \textbf{D}omain knowledg\textbf{E}). PARADE contains 4,778 (46.9\%) paraphrases and 5,404 (53.1\%) non-paraphrases that describe 788 distinct entities from the computer science domain and is the \textit{first publicly available benchmark} for paraphrase identification based on domain knowledge.
	
	\item Third, we evaluate the quality of state-of-the-art paraphrase identification models on PARADE and existing paraphrase identification datasets like MRPC and QQP. We find that both state-of-the-art neural models (which have shown strong performance on existing PI datasets) and non-expert human annotators have poor performance on PARADE. For example, BERT after fine-tuning only achieves 0.709 in terms of F1 on PARADE compared to 0.893 on MRPC and 0.877 on QQP. Such a gap indicates the need for new models that can better exploit specialized domain knowledge.

	
	\item Finally, we show that incorporating external domain knowledge into the training of models like BERT offers the potential for improvements on PARADE. Concretely, we find that SciBERT -- a BERT variant pre-trained on a corpus of computer science papers -- improves the accuracy from 0.729 to 0.741. This improvement is encouraging, and suggests the need for further enhancements in incorporating domain knowledge into NLP models.
	
	
	%
	
	
\end{itemize}

\section{Related Work}
\label{related work}

\noindent\textbf{Framework for Collecting Paraphrases:}
The basic idea of collecting a paraphrase dataset is to connect parallel data that are related to the same reference, like different news articles reporting the same event (MRPC) \cite{dolan-brockett-2005-automatically, dolan-etal-2004-unsupervised}, multiple descriptions of the same video clip \cite{chen2011collecting}, multiple phrasal paraphrases on the web to describe the same concept \cite{hashimoto2011extracting}, different translations of a foreign novel \cite{barzilay-elhadad-2003-sentence}, and multiple tweets that relate to the same topic \cite{xu2014extracting} or contain the same URL \cite{lan2017continuously}. 

In this paper, we propose a novel framework to collect sentential paraphrases from online user-generated flashcards, where different definitions (on the back of flashcards) of the same entity (on the front of flashcards) are probably paraphrases. The main advantage of this framework is that it can easily collect domain-specific paraphrases. Since flashcard websites like Quizlet are mainly used by students to prepare for quizzes and exams, these flashcards are often organized by subject, providing a rich source of domain-specific paraphrases.

\noindent\textbf{Datasets for Paraphrase Identification:}
To our best knowledge, there are five publicly available sentential paraphrase identification datasets: Microsoft Research Paraphrase Corpus (MRPC) \cite{dolan-brockett-2005-automatically, dolan-etal-2004-unsupervised} contains 5,801 pairs of sentences from news articles, PIT-2015 \cite{xu2014extracting} contains 18,762 pairs of tweets on 500 distinct topics, Twitter-URL \cite{lan2017continuously} contains 51,524 pairs of tweets containing 5,187 distinct URLs, Quora Question Pairs (QQP) \cite{iyer2017first} contains 400K\footnote{The size of QQP is much larger than other datasets but its authors claim that the ground-truth labels are not guaranteed to be perfect.} pairs of question pairs on Quora and PAWS \cite{zhang-etal-2019-paws} contains 53,402 pairs of sentences by using word scrambling methods based on QQP. These datasets were not originally designed and constructed from the perspective of domain knowledge. Hence, we present PARADE, the first sentential dataset for paraphrase identification based on domain knowledge as shown in Table \ref{tab:Examples of paraphrases based on domain knowledge}, as a complement to these previous efforts.



\noindent\textbf{Domain-Specific Phrasal Paraphrases:}
Some previous work aims to extract domain-specific \textbf{phrasal} paraphrases \cite{pavlick2015domain, zhang-etal-2016-extract, ma2019essentia}, like ``head'' and ``skull'' in the Biology domain. In this paper, we focus on sentential paraphrases rather than phrasal paraphrases, which require models that consider context and domain knowledge.

\noindent\textbf{Pre-trained Language Models with Domain Knowledge:}
Recently, some works have sought to incorporate domain knowledge into pre-trained language models such as BERT. For example, SciBERT \cite{beltagy2019scibert} uses the same architecture as BERT-base but is pre-trained over a corpus of 1.14M papers, with 18\% of papers from the computer science domain and 82\% from the biomedical domain. It has been reported that SciBERT outperforms BERT-base which is pre-trained over Wikipedia and bookscorpus on a variety of tasks like named entity recognition in the both domains.


\section{Collecting Domain-Specific Paraphrases from Online Flashcards}
\label{ref: section: collecting data framework}
In this section, we propose a novel framework that constructs a domain-specific paraphrase corpus from online user-generated flashcards. We choose \textit{computer science} as the target domain in this paper as a first step. The framework can be easily applied to construct datasets of other domains.


Many web platforms provide flashcards like Quizlet, StudyBlue, AnkiWeb, and CRAM. Each flashcard generated by a user is made up of an entity on the front and a definition describing or explaining the entity on the back. The purpose of flashcards is to help users to understand and remember concepts like ``machine code.''

Our core idea is that two different definitions probably express the same meaning if they have the same entity on the front. Hence, they can be paired as a candidate paraphrase. Our framework can collect arbitrarily many definitions generated by users independently, leading to broad coverage of how native speakers are likely to describe an entity in a specialized domain. By pairing the variety of definitions about concepts (like ``machine code"), paraphrases and non-paraphrases that requires specialized domain (e.g., computer science) knowledge to discern are generated and collected.

\subsection{Collecting Entity-Definition Pairs Related to Specialized Domains}
\label{sub: Collecting Entity-Definition Pairs Related to specialized Domains}
We first collect domain-specific terminology and then collect entity-definition pairs from a popular flashcard website.

\medskip
\noindent\textbf{Domain-specific terminology:}  \citet{ren2014cluscite} presented a dataset of 55,171 research papers in the computer science domain, collected from 2,414 conferences or journals, covering sub-fields like artificial intelligence, computer architecture, networking, and so on. Naturally, high document frequency phrases from these papers can be regarded as computer science terminology. Therefore, 3,813 phrases with document frequency higher than 20 are extracted from these papers, where examples are shown in Table \ref{tab:examples of terminologies}:

\begin{table}[htbp]
  \centering
\small
\setlength{\tabcolsep}{2.0pt}
\renewcommand\arraystretch{0.9}
  \caption{Examples of Computer Science Terminology with Document Frequency (DF)}
    \begin{tabular}{cc|cc}
	\toprule
    Phrases & DF & Phrases & DF \\
	\midrule
    sensor networks & 939   & mobile devices & 425 \\
    information retrieval & 688   & source code & 375 \\
    data structures & 467   & data structure & 348 \\
    query processing & 429   & software systems & 341 \\
	\bottomrule
    \end{tabular}%
  \label{tab:examples of terminologies}%
\end{table}%
	
Next, we use these phrases as queries to search flashcards related to computer science from Quizlet, a well-known online flashcards website with a convenient search  API.\footnote{\url{https://quizlet.com/subject/sensor-networks/}} To ensure paraphrases generated from the flashcards are related to the target domain, we only keep the flashcards where the entity on the front is drawn from our computer science terminology set (of size 3,813). Some example flashcards are presented in Table \ref{tab:examples of flashcards}. 


\begin{table}[htbp]
  \centering
\small

\setlength{\tabcolsep}{0.7pt}
\renewcommand\arraystretch{1.0}
\newcommand{\tabincell}[2]{\begin{tabular}{@{}#1@{}}#2\end{tabular}}
  \caption{Examples of Flashcards Related to Computer Science Domain}
    \begin{tabular}{c|l}
	\toprule
    Entity (Front) & \ Definition (Back) \\
	\midrule
    \tabincell{c}{Artificial  \\Intelligence} & \tabincell{l}{\ s1: simulating logical thoughts, patterns\\ \ and responses} \\
	\hline
    \tabincell{c}{Artificial  \\Intelligence} & \tabincell{l}{\ s2: simulates human thinking and behavior, \\ \ such as the ability to reason and learn} \\
	\hline
    \tabincell{c}{Artificial  \\Intelligence} & \tabincell{l}{\ s3: the ability of a computer or a robot to \\ \ learn from new information} \\
	\hline
	\tabincell{c}{Artificial  \\Intelligence} & \tabincell{l}{\ s4: machines that can apply and acquire \\ \ knowledge}\\
	\bottomrule
    \end{tabular}%
  \label{tab:examples of flashcards}%
\end{table}%

For each flashcard, we extract the entity from the front and the definition from the back to form an entity-definition pair. Since there are many duplicate entities and definitions on Quizlet flashcards, we merge the same definitions and group unique definitions by entities. Further, we only keep entities and definitions in English and in the form of pure text (some definitions contain images) and remove entities with fewer than 5 unique definitions. Finally, 30,917 unique entity-definition pairs are obtained. 


	
\subsection{Generating Candidate Paraphrases}
For the definitions that describe or explain the same entity, it is not guaranteed that any two of them will form a paraphrase because the definitions might focus on different aspects or facets of the entity. An example is shown in Table \ref{tab:examples of flashcards}, where the first two definitions s1 and s2 focus on the aspect of ``simulation'' while the other two definitions s3 and s4 focus on ``learning new knowledge.'' Two definitions on different aspects of the entity are probably not a paraphrase. As a consequence, a random pair of definitions about the same entity has a low probability of expressing the same meaning. 

\begin{table*}[htbp]
  \centering
\small
  \caption{Annotation Criteria}
    \begin{tabular}{p{48.75em}}
\toprule
    \textbf{3- Completely equivalent:} they clearly describe the same computer science concept with same details; \\
    \textbf{Example of label 3:} \\
    Text 1: its software that is freely available and its source code is also available. \\
    Text 2: typically free software where source code is made freely available \\
    Reason: the two sentences are clearly about the same concept (``open source software'') with similar details. \\
\midrule
    \textbf{2 - Mostly equivalent:} as they clearly describe the same computer science concept but some unimportant information differ. Unimportant information include two categories: (1) some examples to explain the entity; and (2) some details can be inferred (based on computer science knowledge) from the overlapping part of the two texts; \\
    \textbf{Example of label 2:} \\
    Text 1: moves packets between computers on different networks. routers operate at this layer. ip and ipx operate at this layer. \\
    Text 2: osi layer that moves packets between computers on different networks. routers \& ip operate at this level. \\
    Reason: they are talking about the same concept: ``network layer'', only some unimportant information differ (the detail ``osi layer'' in Text 2 can be inferred based on computer science knowledge: network layer is one of the layers in OSI model). \\
\midrule
    \textbf{1 - Roughly equivalent:} as they describe the same computer science concept but some important information differs or is missing; Important information here include any details except for the two categories in the previous criterion of label 2; \\
    \textbf{Example of label 1:} \\
    Text 1: term for when a scan fails to find real vulnerabilities. leaves unidentified risk in the code.  \\
    Text 2: malicious activity goes undetected. \\
    Reason: the two sentences might be talking about the same concept: ``false negatives'', but some important information differ (the detail ``...risk in the code...'' in Text 1 cannot be inferred from Text 2 based on computer science knowledge).  \\
\midrule
    \textbf{0 - Not equivalent:} as they describe two different computer science concepts; \\
    \textbf{Example of label 0: }\\
    Text 1: test without knowledge of system internals. \\
    Text 2: attacker has no knowledge of the network environment (external attack).  \\
    Reason: the first sentence is talking about ``system test'' while the second one is about ``system attack.'' \\
\bottomrule
    \end{tabular}%
  \label{tab: Annotation Criteria}%
\end{table*}%

\medskip
\noindent\textbf{Clusters of Definitions:} Hence, we propose to cluster definitions of each entity to group entity-definition pairs that are likely to be on the same aspect. Intuitively, definitions that focus on the same aspect often share some overlapping terms and are likely to be grouped into the same cluster, and pairs of definitions from the same cluster are more likely to be a paraphrase, like s1 and s2, and s3 and s4 in Table \ref{tab:examples of flashcards}. The definitions are first preprocessed with tokenization and lemmatization.\footnote{The tokenizer and lemmatizer are from NLTK.} K-means is applied to cluster the definitions for each entity, where each definition is represented by the average of 300-dimensional word2vec \cite{mikolov2013distributed} token embeddings trained over these definitions. Empirically, we set the number of clusters be half the number of definitions for each entity. Such a large number of clusters is helpful to filter out some noisy data like meaningless or ill-formed definitions because they are likely to be grouped into a single definition's cluster that can be discarded.  

\medskip
\noindent\textbf{Sampling Candidate Paraphrases:} Then, every two of the definitions from the same cluster are paired as a candidate paraphrase. Following \citet{lan2017continuously}, we also filter out paraphrases where the two definitions are very similar like they only differ in punctuation or some typos, or one definition is a sub-string of the other. After that, we collect all candidate paraphrases and obtain a dataset with 10,182 pairs. 

\section{PARADE Dataset}
In Section \ref{ref: section: collecting data framework}, we introduced our framework for generating domain-specific candidate paraphrases. Although each one of the candidate paraphrases is focused on the same topic (entity), we still need to confirm that the two definitions  express the same meaning. In this section, we introduce our annotation strategy for candidate paraphrases and formally present the PARADE dataset for paraphrase identification based on domain knowledge.

\subsection{Annotators with Domain Expertise}
As discussed in Section \ref{section: Introduction}, candidate paraphrases in our dataset can not be annotated correctly without specialized domain knowledge. Hence, unlike most previous works \cite{lan2017continuously, xu2015semeval, xu2014extracting, chen2011collecting} that hire workers from crowdsourcing platforms like Amazon Mechanical Turk, we invited students majoring in computer science as the annotators for this dataset. The 40 invited annotators include 5 Ph.D. students, 18 masters students, and 17 upper-level undergraduates. All have finished courses that cover almost all of the entities (topics) introduced in Section \ref{sub: Collecting Entity-Definition Pairs Related to specialized Domains}.

\subsection{Annotation Criteria}
Since the annotators have domain expertise, we expect them to provide more specific judgments than just true paraphrase or not. The annotation criteria are presented in Table \ref{tab: Annotation Criteria}: Completely equivalent (3), Mostly equivalent (2), Roughly equivalent (1), and Not equivalent (0). Labels of 3 and 2 are considered paraphrases, while 0 and 1 are  non-paraphrases.

\subsection{Annotation Quality Control}


Annotators are asked to carefully read the annotation criteria before starting annotations. Each pair is randomly assigned to three annotators; the final ground-truth is decided by majority vote. We evaluate annotation quality of each annotator via Cohen's Kappa score \cite{artstein2008inter} against the ground-truth. 
The average Cohen's Kappa score of the annotators is 0.65. 
Following \citet{lan2017continuously}, we re-assign the data instances that were assigned to 2 annotators with low annotation quality (Cohen's Kappa score$<$0.4) to the best 5 annotators (Cohen's Kappa score$>$0.75) and ask them to re-label (give labels without seeing old labels) these data instances. 

\subsection{Dataset Description}
Finally, we construct the first gold dataset for paraphrase identification based on domain knowledge, with 10,182 pairs of definitions that describe 788 distinct entities in the computer science domain. Among them, 4,778 (46.9\%) are paraphrases and 5,404 (53.1\%) are non-paraphrases. The average length of the definitions is 17.1 words and the maximum length is 30. An example from PARADE is shown in Table \ref{tab:An example of PARADE}. Note that entities like ``machine code'' are also provided with definitions. However, these entities are not used in training and testing models for paraphrase identification tasks; otherwise the models will just learn the answers.

\begin{table}[htbp]
  \centering
\small
\setlength{\tabcolsep}{0.7pt}
\renewcommand\arraystretch{1.0}
\newcommand{\tabincell}[2]{\begin{tabular}{@{}#1@{}}#2\end{tabular}}
  \caption{An example of PARADE}
    \begin{tabular}{l}
 \toprule
 	Entity: Machine Code\\
    \tabincell{l}{Definition 1:	the lowest level of code made up of 0s and 1s.} \\
    \tabincell{l}{Definition 2:	binary instructions used by the cpu.} \\
    Label: paraphrase \\
 \bottomrule
    \end{tabular}%
  \label{tab:An example of PARADE}%
\end{table}%

\begin{figure}[t]
    \centering
    \includegraphics[scale=0.35]{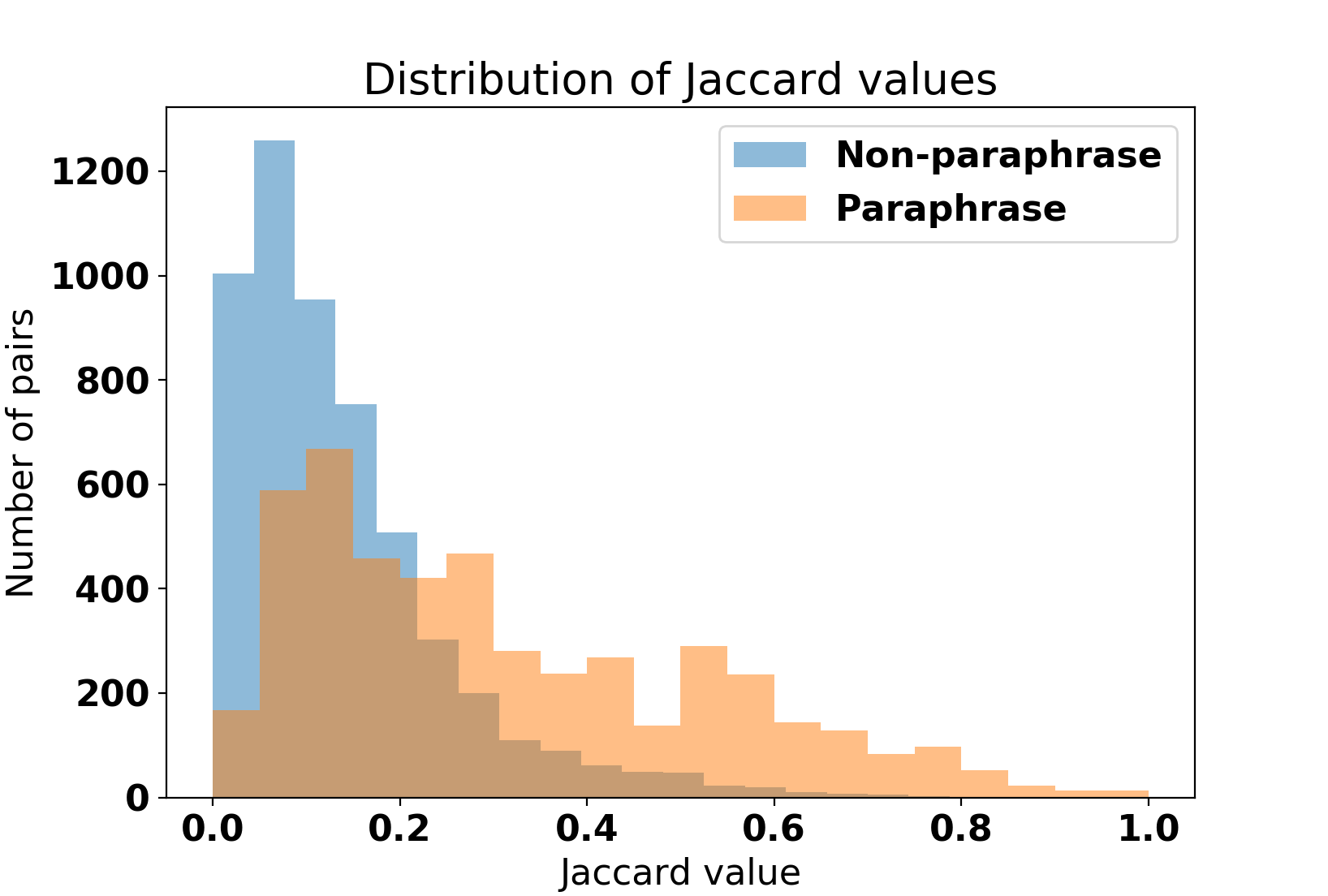}
    \caption{Distributions of Jaccard similarity for paraphrases and non-paraphrases in PARADE.}
    \label{fig: Distributions of Jaccard similarity for paraphrases and non-paraphrases in PARADE.}
\end{figure}

We calculate the Jaccard similarity for each pair to measure the lexical overlap\footnote{Stopwords and punctuation were removed; words were stemmed.} between the two definitions. In Figure \ref{fig: Distributions of Jaccard similarity for paraphrases and non-paraphrases in PARADE.}, we illustrate the distributions of Jaccard similarity for paraphrases and non-paraphrases. It can be observed that PARADE contains many paraphrases that overlap very little at the lexical level but are semantically equivalent. In addition, PARADE contains a few non-paraphrases that overlap clearly but are not semantically equivalent.

In Section \ref{experiments}, we present a qualitative analysis of PARADE on the cases where BERT give wrong predictions, which indicates that PARADE is truly enriched with domain knowledge.

\section{Experiments}
\label{experiments}

In this section, we present experiments that aim to answer the following research questions (RQs):
\begin{itemize}
	\item RQ1: How do BERT and other neural models perform on PARADE? Better or worse than their performance on traditional PI datasets?
	\item RQ2: What kinds of domain knowledge are captured by PARADE? And how well do non-experts identify paraphrases that contain this domain knowledge?
	\item RQ3: Can we achieve high-quality identification by augmenting BERT-like models with a collection of domain-specific resources? 
\end{itemize}

\subsection{Experimental Setup}
We first introduce our experimental setup here, including paraphrase identification models, other PI datasets and their partition and reproducibility.

\medskip
\noindent\textbf{Models for Binary Paraphrase Identification: } We test seven different approaches on PARADE. The Decomposable Attention Model (\textbf{DecAtt},  380K parameters) \cite{parikh-etal-2016-decomposable} is one of the earliest models to apply attention for modeling sentence pairs. It computes the word pair interaction between the two sentences in a candidate paraphrase. The Pairwise Word Interaction Model (\textbf{PWIM}, 2.2M parameters) \cite{he2016pairwise} uses Bi-LSTM to model the context of each word and then uses cosine similarity, Euclidean distance and dot product together to model word pair interactions. The Enhanced Sequential Inference Model (\textbf{ESIM},  7.7M parameters) \cite{chen-etal-2017-enhanced} first encodes sentences by using Bi-LSTM and then also calculates the word pair interaction between the two sentences like DecAtt. The Shortcut-Stacked Sentence Encoder (\textbf{SSE}, 140M parameters)  \cite{nie2017shortcut} applies a stacked Bi-LSTM with skip connections as the sentence encoder. Recently, the Bidirectional Encoder Representations from Transformer (\textbf{BERT}) \cite{devlin2018bert} obtains the state-of-the-art performance on many NLP tasks, including paraphrase identification. We evaluate \textbf{BERT-base} (12 layers and 768 hidden embedding size with 108M parameters) and \textbf{BERT-large} (24 layers and 1024 hidden embedding size with 334M parameters) on PARADE. We also adopt \textbf{ALBERT}, which compresses the architecture of BERT by factorized embedding parameterization and cross-layer parameter sharing, to obtain a substantially higher capacity than BERT. We choose the maximum version ALBERT-xxlarge (12 layers and 4096 hidden embedding size with 235M parameters).


\medskip
\noindent\textbf{Datasets and Their Partition:} For PARADE, we randomly split it by entities into three parts: 7,550 with 560 distinct entities in the training set, 1,275 with 110 distinct entities in the validation set  and 1,357 with 118 distinct entities in the testing set. For paraphrase datasets MRPC\footnote{\url{https://gluebenchmark.com/tasks}}, PAWS\footnote{\url{https://github.com/google-research-datasets/paws}}, Twitter-URL\footnote{\url{https://github.com/lanwuwei/Twitter-URL-Corpus}} and PIT-2015\footnote{\url{https://cocoxu.github.io/##publications}}, we follow the data partitioning strategy of their authors. For QQP\footnote{\url{https://gluebenchmark.com/tasks}}, the labels for its test set at GLUE are private, so we treat its validation set at GLUE as the test set and sample another part from its training set as the validation set. Details of these previous PI datasets can be found in Section \ref{related work}. 

\medskip
\noindent\textbf{Reproducibility: } PARADE and its split in this paper is released.\footnote{\url{https://github.com/heyunh2015/PARADE_dataset}} For BERT, we use a widely used pytorch implementation\footnote{\url{https://github.com/huggingface/transformers}} and Adam optimizer with batch size 32 and learning rate 2e-5. We fine-tuned BERT for 20 epochs. We selected the BERRT hyper-parameters from the range as recommended in \citet{devlin2018bert} and based on the performance in terms of F1 on the validation set. The implementations\footnote{\url{https://github.com/lanwuwei/SPM\_toolkit}} of the other neural models are from \citet{lan-xu-2018-neural}, and we use the same hyper-parameters as recommended by \citet{lan-xu-2018-neural}. 


\begin{table}[htbp]
  \centering
 \small

\setlength{\tabcolsep}{0.7pt}
\renewcommand\arraystretch{1.0}
  \caption{Performance of BERT on paraphrase identification datasets}
    \begin{tabular}{lcccc}
\toprule
    BERT-large    & \multicolumn{1}{c}{Accuracy} & \multicolumn{1}{c}{F1} & \multicolumn{1}{c}{Precision} & \multicolumn{1}{c}{Recall} \\
\midrule
    MRPC  & 0.853 &	0.893 & 0.866 & 0.922\\
    QQP   & 0.908 & 0.877 & 0.866 & 0.889 \\
    PWAS  & 0.939 & 0.933 & 0.923 & 0.944 \\
    Twitter-URL & 0.905 & 0.770 & 0.728 & 0.817 \\
    PIT-2015 & 0.901 &	0.746 &	0.803	& 0.697 \\
    PARADE & 0.736 &	0.709 & 0.669	& 0.753\\
\bottomrule
    \end{tabular}%
  \label{tab:Performance of BERT on paraphrase identification datasets}%
\end{table}%

\begin{table}[htbp]
  \centering
\small
\setlength{\tabcolsep}{0.7pt}
\renewcommand\arraystretch{1.0}
  \caption{Performance of Neural Models on PARADE}

    \begin{tabular}{lcccc}
\toprule
          & Accuracy & F1    & Precision & Recall \\
\midrule
    DecAtt & 0.540 & 0.530 & 0.519 & 0.541 \\
    ESIM  & 0.595	& 0.646	& 0.556	& 0.770 \\
    PWIM  & 0.701 & 0.687 & 0.689 & 0.686\\
    SSE   & 0.689	& 0.702	& 0.649	& 0.764 \\
    BERT-base & 0.729 & 0.708  &  0.687  & 0.731 \\
    BERT-large & 0.736 &	0.709 & 0.669	& 0.753\\
    ALBERT-xxlarge & 0.753 &	 0.741 & 0.738	& 0.745 \\
\bottomrule
    \end{tabular}%
  \label{tab:Performance of Neural Models on PARADE}%
\end{table}%

\subsection{RQ1: Paraphrase Identification Comparison}
\label{sec: Performance Comparison}

We first present the performance of BERT-large on PARADE and previous PI datasets in Table \ref{tab:Performance of BERT on paraphrase identification datasets}. Compared to datasets that lack domain knowledge, we observe that BERT yields the lowest performance on PARADE across all metrics. For example, BERT obtains 0.709 in terms of F1, which is much lower than its performance on the other datasets. 
Both the precision and the recall are relatively low, which indicates that identifying paraphrases in PARADE is non-trivial even for BERT. 

Additionally, we present the results of BERT-base, BERT-large, ALBERT-xxlarge and other neural models on PARADE in Table \ref{tab:Performance of Neural Models on PARADE}. We observe that the other neural models have lower performance than BERT-family models. Among the BERT-family models, BERT-large is slightly better than BERT-base, and ALBERT-xxlarge is 
the best due to its large learning capacity.
However, the best performance is still relatively low on this dataset.    

A possible reason is that these general neural net models do not sufficiently capture specialized knowledge of the computer science domain. BERT is pre-trained on two corpora: BooksCorpus (800M words) \cite{zhu2015aligning} and English Wikipedia (2,500M words), which leads to some world knowledge learned as reported in \citet{petroni-etal-2019-language, davison-etal-2019-commonsense}. However, BooksCorpus\footnote{This corpus has 11,038 books like \textit{American Psycho} and \textit{No Country for Old Men.}} does not contain computer science books. While Wikipedia does contain articles on computer science, BERT may not pay enough attention to this subject since Wikipedia is such a huge corpus and computer science is just one branch. 

\subsection{RQ2: Domain Knowledge}
\label{section: RQ2: Domain Knowledge}

As discussed in Section \ref{sec: Performance Comparison}, BERT and other neural models face key challenges in paraphrase identification with domain knowledge as in PARADE. A possible reason is that PARADE has a lot of domain knowledge, which is beyond the lexical, syntactic features, or even commonsense knowledge captured by these models. To confirm the presence of domain knowledge, we first conduct a qualitative analysis of PARADE. 

\begin{table}[htbp]
  \centering
 \small
\setlength{\tabcolsep}{0.7pt}
\renewcommand\arraystretch{1.0}
\newcommand{\tabincell}[2]{\begin{tabular}{@{}#1@{}}#2\end{tabular}}
\caption{A case where BERT predict incorrectly}
    \begin{tabular}{ll}
\toprule
	Entity: Type Inference\\
    \tabincell{l}{Definition 1: variables don't need explicit statements about \\their type unlike in \textbf{java}. \textbf{haskell} can automatically tell that\\ \textbf{1} is of type \textbf{int}. } &  \\
   \tabincell{l}{Definition 2: allows the \textbf{compiler} to deduce the proper type \\for you automatically, instead of you having to say it. }&  \\
\midrule
    BERT Prediction: non-paraphrase &  \\
    Ground-truth: paraphrase &  \\
\bottomrule
    \end{tabular}%
  \label{tab:An example of cases where BERT predict incorrectly}%
\end{table}%

\medskip
\noindent\textbf{Qualitative Analysis: } We qualitatively analyzed 277 cases (171 paraphrase and 106 non-paraphrases) where BERT predicts the wrong results. From the perspective of domain knowledge, we count the occurrences of each phenomenon in the following categories: \textbf{Specialized Terminology.} Examples in the computer science domain include java, haskell and compiler in Table \ref{tab:An example of cases where BERT predict incorrectly}. \textbf{Acronyms and Abbreviations:} Examples include ``int'' for integer in Table \ref{tab:An example of cases where BERT predict incorrectly}, OS (operating system), OSI (Open Systems Interconnection) model and so on. \textbf{Numbers and Equations:} These have special meaning like ``port: 80'', ``arr[i][j] == 1'' and ``an m-ary tree with m = 2''. \textbf{Inference:} These non-overlapping sentences may be paraphrases based on domain-specific inference. For example in Table \ref{tab:An example of cases where BERT predict incorrectly}, definition 1 does not mention ``compiler'' in definition 2 but domain experts can infer that based on context and domain knowledge that the compiler is responsible for identifying the type. \textbf{Examples:} Non-overlapping paraphrases use examples to support the main idea, like the example of ``haskell'' in the definition 1 in Table \ref{tab:An example of cases where BERT predict incorrectly}. Although definition 2 does not have this example, they still express the concept of ``type inference.''


A typical example of the cases where these phenomena occur together is shown in Table \ref{tab:An example of cases where BERT predict incorrectly}. We report the number of occurrences of each phenomenon in Table \ref{tab:Number of domain knowledge phenomenons} and observe that the cases where BERT fails have a high frequency of these domain knowledge phenomena, further supporting the assertion that PARADE is enriched with domain knowledge.

\begin{table}[htbp]
  \centering
 \small
\setlength{\tabcolsep}{0.7pt}
\renewcommand\arraystretch{1.0}
  \caption{Number of domain knowledge phenomena in the 277 cases where BERT mis-labels}
    \begin{tabular}{lcc}
\toprule
   \multicolumn{1}{l}{Phenomenon}       & \multicolumn{1}{l}{Count} & \multicolumn{1}{l}{Frequency} \\
\midrule
    Specialized Terminology & 150   & 0.54 \\
    Acronyms and Abbreviations & 30    & 0.11 \\
    Numbers and Equations & 31    & 0.11 \\
    Inference & 114   & 0.41 \\
    Examples & 78    & 0.28 \\
    Cases that have one phenomenon at least & 197  & 0.71  \\
\bottomrule
    \end{tabular}%
  \label{tab:Number of domain knowledge phenomenons}%
\end{table}%

\begin{table}[htbp]
  \centering
 \small
\setlength{\tabcolsep}{0.7pt}
\renewcommand\arraystretch{1.0}
  \caption{Performance of Non-Experts on Paraphrase Identification}
    \begin{tabular}{lcccc}
\toprule
     Human     & Accuracy & F1    & Precision & Recall \\
\midrule 
    MRPC  & 0.70  & 0.75  & 0.77  & 0.74 \\
    QQP   & 0.74  & 0.61  & 0.50  & 0.77 \\
    PAWS  & 0.90  & 0.88  & 0.86  & 0.90 \\
    Twitter-URL & 0.90  & 0.71  & 0.67  & 0.75 \\
    PIT-2015 & 0.90	& 0.76	& 0.80	& 0.73 \\
    PARADE & 0.62  & 0.56  & 0.45  & 0.73 \\
\bottomrule
    \end{tabular}%
  \label{tab:Performance of Human Beings without Computer Science Domain Knowledge}%
\end{table}%

\medskip
\noindent\textbf{Performance of Non-Experts without Domain Knowledge:}
To further confirm the presence of domain knowledge, we invite three college students who are \textbf{not} majoring in computer science to label PARADE and other datasets. Before evaluation, we ask the students to carefully read the annotation criteria of each dataset and 100 sampled cases with labels from the training set from each dataset. After that, 100 cases without ground-truth are randomly sampled from the test set of each dataset for evaluating the quality of non-expert annotators. 

The results are presented in Table \ref{tab:Performance of Human Beings without Computer Science Domain Knowledge}. We observe that non-experts without domain knowledge obtain abysmal performance on PARADE like 0.56 in terms of F1. However, on other datasets, these non-experts can achieve much better results like 0.88 in terms of F1 on PAWS. By interviewing these students, we believe they can correctly identify paraphrases based on lexical, syntactic and commonsense knowledge on all datasets except for PARADE, where the lack of specialized domain knowledge made the task too challenging.

\subsection{RQ3: Incorporating Domain Knowledge}
As shown in Section \ref{sec: Performance Comparison} and Section \ref{section: RQ2: Domain Knowledge}, both 
widely used neural models and non-expert human annotators have poor performance on PARADE. 
To corroborate the importance and possibility to enhance a model for PARADE by incorporating specialized domain knowledge,  we ran an off-the-shelf model,  SciBERT\footnote{\url{https://huggingface.co/allenai/scibert\_scivocab\_uncased}} \cite{beltagy2019scibert}, that uses the same architecture as BERT-base and is pre-trained using 1.14M 
papers from Semantic Scholar  \cite{ammar-etal-2018-construction} with 18\% of papers from the computer science domain and 82\% from the biomedical domain. 
As shown in Table \ref{tab:Results of Enhancing BERT by incorporating domain knowledge}, 
SciBERT outperforms BERT consistently over all the metrics. This experiment shows that simply 
using corpora of 
a target domain for model training 
does lead to some improvements on PARADE. 
Further improvements may be achieved by methods that can more effectively infuse domain knowledge into NLP models. 

\begin{table}[htbp]
  \centering
 \small
\setlength{\tabcolsep}{0.7pt}
\renewcommand\arraystretch{1.0}
  \caption{Results of Enhancing BERT by incorporating domain knowledge}
    \begin{tabular}{lcccc}
\toprule
         & Accuracy & F1    & Precision & Recall \\
\midrule
    BERT-base & 0.729 & 0.708 & 0.687	& 0.731 \\
    SciBERT & 0.741$\uparrow$ & 0.723$\uparrow$ & 0.707$\uparrow$ & 0.740$\uparrow$ \\
\bottomrule
    \end{tabular}%
  \label{tab:Results of Enhancing BERT by incorporating domain knowledge}%
\end{table}%

\section{Conclusion and Future Work}
We have presented PARADE, a new dataset for sentential paraphrase identification requiring domain knowledge. 
We conducted extensive experiments and analysis showing that both  state-of-the-art neural models and non-expert human annotators
perform poorly on PARADE. 
In the future, we will continue to investigate effective ways to obtain domain knowledge and incorporate it into enhanced models for paraphrase identification. In addition, since PARADE provides entities like ``machine code'' for definitions, this new dataset could also be useful for other tasks like entity linking \cite{shen2014entity}, entity retrieval \cite{petkova2007proximity} and entity or word sense disambiguation \cite{navigli2009word}.
%
\section*{Acknowledgments}
This work is supported in part by NSF (\#IIS-1909252).

\bibliography{acl2020}
\bibliographystyle{acl_natbib}



\end{document}